\documentclass{article}       
\usepackage{cite}
\usepackage[height=8.5in,width=6.2in]{geometry}

\usepackage{graphicx}
\usepackage{times}
\usepackage{xspace}
\usepackage{enumerate}
\usepackage{amsmath,amssymb,amsthm,bm}
\usepackage{array}
\usepackage[algo2e,boxed,vlined]{algorithm2e}

\newcommand{\pol}[1]{{#1}^{\circ}}

\newcommand{\tb}[1]{{\bf #1}}
\newcommand{\proj}{\text{proj}}

\newcommand{\mtl}{\textsc{Mtl}\xspace}
\newcommand{\spgfp}{SPG$_{\text{FP}}$\xspace}
\newcommand{\spgqp}{SPG$_{\text{QP}}$\xspace}

\newcommand{\mynorm}[2]{\| {#1} \|_{#2}}
\newcommand{\norm}[1]{\mynorm{#1}{}}
\newcommand{\pnorm}[2]{\mynorm{#1}{#2}}

\newcommand{\norml}[1]{\mynorm{#1}{1}}

\newcommand{\infnorm}[1]{\mynorm{#1}{\infty}}

\newcommand{\oneinf}{\ell_{1,\infty}}

\newcommand{\oneinfnorm}[1]{\mynorm{#1}{1,\infty}}

\newcommand{\enorm}[1]{\mynorm{#1}{2}}

\newcommand{\frob}[1]{\|{#1}\|_{\text{F}}}

\newcommand{\ip}[2]{\langle {#1},\, {#2} \rangle}

\newcommand{\nlsum}{\sum\nolimits}
\newcommand{\nlmin}{\min\nolimits}

\newcommand{\reals}{\mathbb{R}}
\newcommand{\half}{\tfrac{1}{2}}

\newcommand{\vy}{\bm{y}}

\newcommand{\tx}{\mathsf{X}}

\newcommand{\ty}{\mathsf{Y}}

\newcommand{\mx}{\bm{X}}

\newcommand{\sml}[1]{{\small #1}}
\newcommand{\Lc}{\mathcal{L}}
\newcommand{\fromto}[3]{\sml{$#1 \le #2 \le #3$}}
\newcommand{\set}[1]{\left\{ {#1}\right\}}

\DeclareMathOperator*{\argmin}{argmin}
\DeclareMathOperator{\sgn}{sgn}
\DeclareMathOperator{\trace}{tr}
\DeclareMathOperator{\prox}{prox}
\DeclareMathOperator{\Diag}{Dg}

\DeclareMathOperator{\vect}{vec}

\newtheorem{theorem}{Theorem}
\newtheorem{lemma}[theorem]{Lemma}

\theoremstyle{definition}
\newtheorem{definition}[theorem]{Definition}

\numberwithin{equation}{section}

\begin{document}

\title{Fast projections onto mixed-norm balls with applications\thanks{Preprint of a paper under review}}

\author{Suvrit Sra\\
  MPI for Intelligent Systems, T\"ubingen, Germany\\
  Tel.: +49-7071-601572\\
  \textit{suvrit@tuebingen.mpg.de}}

\date{Submitted: Sep., 2011}

\maketitle

\begin{abstract}
  Joint sparsity offers powerful structural cues for feature selection,
  especially for variables that are expected to demonstrate a ``grouped''
  behavior. Such behavior is commonly modeled via group-lasso, multitask
  lasso, and related methods where feature selection is effected via
  mixed-norms. Several mixed-norm based sparse models have received
  substantial attention, and for some cases efficient algorithms are also
  available. Surprisingly, several \emph{constrained} sparse models seem to be
  lacking scalable algorithms. We address this deficiency by presenting batch
  and online (stochastic-gradient) optimization methods, both of which rely on
  efficient projections onto mixed-norm balls. We illustrate our methods by
  applying them to the multitask lasso. We conclude by mentioning some open
  problems.

\end{abstract}

\begin{quote}
\end{quote}
\section{Introduction}
\vspace*{-8pt}
Sparsity encodes key structural information about data and permits estimating
unknown, high-dimensional vectors robustly. No wonder, sparsity has been
intensively studied in signal processing, machine learning, and statistics,
and widely applied to many tasks therein. But the associated literature has
grown too large to be summarized here; so we refer the reader
to~\cite{icmltut,bach11,rice,bach2} as starting points.

Sparsity constrained problems are often cast as instances of the following high-level optimization problem
\begin{equation}
  \label{eq.8}
  \nlmin_{x \in \reals^d}\quad L(x) + \lambda f(x), 
\end{equation}
where $L$ is a differentiable loss-function, $f$ is a convex (nonsmooth)
regularizer, and $\lambda > 0$ is a scalar. Alternatively, one may
prefer the constrained formulation
\begin{equation}
  \label{eq.9}
  \nlmin_{x \in \reals^d}\quad L(x)\quad\text{s.t.}\quad f(x) \le \gamma.
\end{equation}

Both formulations~\eqref{eq.8} and~\eqref{eq.9} continue to be actively
researched, the former perhaps more than the latter. We focus on the latter,
primarily because it often admits simple but effective first-order
optimization algorithms. Additional benefits that make this constrained
formulation attractive include:
\begin{itemize}
\item Even when the loss $L$ is nonconvex, gradient-projection remains applicable;
\item If the loss is separable, it is easy to derive highly scalable
  incremental or stochastic-gradient based optimization algorithms;
\item If only inexact projections onto $f(x) \le \gamma$ are possible (a
  realistic case), convergence analysis of gradient-projection-type methods
  remains relatively simple.
\end{itemize}

In this paper, we study a particular subclass of~\eqref{eq.9} that has
recently become important, namely, \emph{groupwise sparse regression}. Two
leading examples are multitask
learning~\cite{evgPon04,evMiPo05,liPaZh09,obTaJo06} and
group-lasso~\cite{yuLi04,tuVeWr05,bach08}. A key component of these regression
problems is the regularizer $f(x)$, which is designed to enforce `groupwise
variable selection'---for example, with $f(x)$ chosen to be a
\emph{mixed-norm}.

\begin{definition}[Mixed-norm]
\label{def.mn}
Let $x \in \reals^d$ be \emph{partitioned} into subvectors
$x^i \in \reals^{d_i}$, for $i \in [m]$\footnote{We use $[m]$ as a shorthand for the set $\set{1,2,\ldots,m}$.}. The $\ell_{p,q}$-\emph{mixed-norm}
for $p$, $q\ge 1$, is then defined as
\begin{equation}
  \label{eq:2}
  f(x) = \pnorm{x}{p,q} := \bigl(\nlsum_{i=1}^m\pnorm{x^i}{q}^p\bigr)^{1/p}.
\end{equation}
\end{definition}

The most practical instances of~\eqref{eq:2} are $\ell_{1,q}$-norms, especially
for $q \in \set{1,2,\infty}$. The choice $q=1$ yields the ordinary
$\ell_1$-norm penalty; $q=2$ is used in group-lasso~\cite{yuLi04}, while
$q=\infty$ arises in compressed sensing~\cite{tropp06} and multitask
lasso~\cite{liPaZh09}. Less common, though potentially useful versions allow interpolating between these extremes by letting $q \in (1,\infty)$; see also~\cite{alain,pmtl,kowal09}. 

Definition~\ref{def.mn} can be substantially generalized: we may allow the
subvectors $x^i$ to overlap; or to even be normed
differently~\cite{zhRoYu09}. But unless the overlapping has special
structure~\cite{jeMaObBa10,bach2,liu.nips10}, it leads to somewhat impractical
mixed-norms, as the corresponding optimization problem~\eqref{eq.9} becomes
much harder. Since our chief aim is to develop fast, scalable algorithms
for~\eqref{eq.9}, we limit our discussion to $\ell_{1,q}$-norms---this choice
is widely applicable, hence
important~\cite{tuVeWr05,liPaZh09,frHaTi10,icml10,beScFrMu08,simila,fobos,obTaJo06,evMiPo05}.

Before moving onto the technical part, we briefly list the paper's main
contents:\footnote{Which also helps position this paper relative to its
  precursor at ECML 2011~\cite{ecml11}.}
\begin{itemize}
\item Batch and online (stochastic-gradient based) algorithms for
  solving~\eqref{eq.9};
\item Theory of and algorithms for fast projection onto $\ell_{1,q}$-norm balls;
\item Application to $\ell_{1,q}$-norm based multitask lasso; both batch and online versions;
\item Application to computing projections for matrix mixed-norms;
\item A set of open problems.
\end{itemize}

\vspace*{-12pt}
\section{Basic theory}
\vspace*{-6pt}
We begin by developing some basic theory. Our aim is to efficiently implement
a generic `first-order' algorithm: Generate a sequence
$\set{x_t}$ by iterating
\begin{equation}
  \label{eq.5}
  x_{t+1} = \proj_{f}(x_t-\eta_t\nabla_t),\quad t=0,1,\ldots,
\end{equation}
where $\eta > 0$ is a stepsize, $\nabla_t$ is an estimate of the gradient, and
$\proj_f$ is the projection operator that enforces the constraint $f(x) \le
\gamma$. Below we expand on the most challenging component of
iteration~\eqref{eq.5} when applied to mixed-norm regression, namely efficient
computation of the projection operator $\proj_f$.

\subsection{Efficient projection via proximity}
\label{sec:proj}
Formally, the (orthogonal) \emph{projection operator} $\proj_f: \reals^d \to \reals^d$ is defined as
\begin{equation}
  \label{eq.proj}
  \proj_{f}(y):= \argmin\nolimits_{x}\quad 
  \half\enorm{x-y}^2\quad\text{s.t.}\quad f(x) \le \gamma.
\end{equation}
Closely tied to projection is the \emph{proximity operator} $\prox_h: \reals^d \times \reals_+ \to \reals^d$
\begin{equation}
  \label{eq.prox}
  \hskip-1.2cm\prox_h(y, \theta) := \argmin\nolimits_{x}\quad\half\enorm{x-y}^2 + 
  \theta h(x),
\end{equation}
where $h$ is a convex function on $\reals^d$. Operator~\eqref{eq.prox}
generalizes projections: if in~\eqref{eq.prox} the function $h$ is chosen to
be the indicator function for the set $\set{x : f(x) \le \gamma}$, then the
operator $\prox_h$ reduces to the projection operator
$\proj_f$. 

Alternatively, for convex $f$ and $h$, operators $\proj_f$ and $\prox_h$ are
also intimately connected by duality. Indeed, this connection proves key to
computing a projection efficiently whenever its corresponding proximity
operator is `easier'. The idea is simple (see e.g.,~\cite{patrik05}), but
exploiting it effectively requires some care; let us see how.

Let $\Lc(x,\theta)$ be the Lagrangian for~\eqref{eq.proj}; and let the optimal dual solution be denoted by $\theta^*$. Assuming strong-duality, the optimal primal solution is given by
\begin{equation}
  \label{eq:14}
  x(\theta^*) := \argmin\nolimits_{x}\ \Lc(x,\theta^*) 
  := \argmin\nolimits_{x}\ \half\enorm{x-y}^2 + \theta^*(f(x) - \gamma).
\end{equation}
But to compute~\eqref{eq:14}, we require the optimal $\theta^*$---the key insight on obtaining $\theta^*$ is that it can be computed by solving a single
nonlinear equation. Here is how.

First, observe that if $f(y) \le \gamma$, then $x(\theta^*)=y$, and there is
nothing to compute. Thus, assume that $f(y) > \gamma$; then, the optimal point
$x(\theta^*)$ satisfies
\begin{equation}
  \label{eq.15}
  f(x(\theta^*))=\gamma.
\end{equation}
Next, observe from~\eqref{eq:14} that for a fixed $\theta$, the point $x(\theta)$ equals the operator $\prox_f(y, \theta)$. Consider, therefore, the nonlinear function (residual)
\begin{equation}
  \label{eq:18}
  g(\theta) := f(x(\theta)) - \gamma\ =\  f(\prox_f(y, \theta))-\gamma,
\end{equation}
which measures how accurately equation~\eqref{eq.15} is satisfied. The optimal $\theta^*$ can be then obtained by solving $g(\theta)=0$, for which the following  lemma proves very useful. 

\begin{lemma}
  \label{lemm:mono}
  Let $f(x)$ be a gauge\footnote{That is, $f$ is nonnegative, positively
    homogeneous, and disappears at the origin~\cite[\S15]{rock70}}, and let
  $g(\theta)$ be as defined in~\eqref{eq:18}. Then, there exists an interval $[0,
  \theta_{\max}]$, on which $g(\theta)$ is monotonically decreasing, and
  differs in sign at the endpoints.
\end{lemma}
\begin{proof}
  By assumption on $f(y)$, it holds that $g(0) = f(y)-\gamma > 0$. We claim
  that for $\theta \ge \pol{f}(y)$, where $\pol{f}$ denotes the \emph{polar}
  of $f$, the optimal point $x(\theta) = 0$. To see why, suppose that $\theta
  \ge \pol{f}(y)$, but $x(\theta) \neq 0$. Then, $\half\enorm{x(\theta)-y}^2 +
  \theta f(x(\theta)) < \half\enorm{y}^2$. But since $\enorm{\cdot}^2$ is
  strictly convex, the inequality $\enorm{y}^2 - \enorm{x-y}^2 < 2\ip{y}{x}$
  also holds for any $x$. Thus, it follows that $\theta <
  \ip{y}{x(\theta)}/f(x(\theta))$, whereby, for $\theta \ge \sup_{x \neq 0}
  {\ip{y}{x}}/{f(x)}=\pol{f}(y)$, the optimal $x(\theta)$ must equal
  $0$. Hence, we may select $\theta_{\max} = \pol{f}(y)$. Monotonicity of $g$
  follows easily, as it is the derivative of the concave (dual) function
  $\inf_{x}\Lc(x,\theta)$. Finally, $g(\theta_{\max}) = -\gamma
  < 0$, so it differs in sign.\qed
\end{proof}

Since $g(\theta)$ is continuous, changes sign, and is monotonic in the interval $[0,\theta_{\max}]$, it has a unique root therein. This root can be computed to $\epsilon$-accuracy using bisection in $O(\log (\theta_{\max}/\epsilon))$ iterations. We recommend not to use mere bisection, but rather to invoke a more powerful root-finder that combines bisection, inverse quadratic interpolation, and the secant method (e.g., \textsc{Matlab}'s \texttt{fzero} function). Pseudocode encapsulating these ideas is given in Algorithm~\ref{algo2}. 

\begin{algorithm2e}[tbp]
  \KwIn{Subroutine to compute $\prox_f(y,\theta)$; vector $y$; scalar $\gamma > 0$}
  \KwOut{$x^* := \proj_f(y,\gamma)$}
  \eIf{$f(y) \le \gamma$}{\KwRet{$x^*=y$}}{
    Define $g(\theta) := \prox_f(y,\theta)-\gamma$\;
    Compute interval $[\theta_{\min},\theta_{\max}] = [0, \pol{f}(y)]$\;
    Compute root $\theta^*  = \textsc{FindRoot}(g(\theta), \theta_{\min},\theta_{\max})$\;
  }
 \KwRet{$x^* = \prox_f(y,\theta^*)$}
 \caption{\small Root-finding for projection via proximity} \label{algo2}
\end{algorithm2e}

\subsubsection{Projection onto $\ell_{1,q}$-norm balls}
After the generic approach above, let us specialize to projections for the
case of central interest to us, namely, $\proj_f$ with $f(x)
=\ell_{1,q}(x)$. Algorithm~\ref{algo2} requires computing the upper bound
$\theta_{\max} = \pol{f}(y)$. To that end Lemma~\ref{lem.dual}, which actually
proves much more, proves useful.

\begin{lemma}[Dual-norm]
  \label{lem.dual}
  Let $p,q \ge 1$; and let $p^*,q^* \ge 1$ be ``conjugate'' scalars, i.e.,
  $1/p+1/p^*=1$ and $1/q+1/q^*=1$. The polar (dual-norm) of
  $\pnorm{\cdot}{p,q}$ is $\pnorm{\cdot}{p^*,q^*}$.
\end{lemma}
\begin{proof}
  By definition, the norm dual to an arbitrary norm $\norm{\cdot}{}$ is given by
  \begin{equation}
    \label{eq.21}
    \pnorm{u}{*} := \sup\set{\ip{x}{u}\ |\ \ \norm{x}{} \le 1}.
  \end{equation}
  To prove the lemma, we prove two items: (i) for any two (conformally partitioned) vectors $x$ and $u$, we have $|\ip{x}{u}| \le \pnorm{x}{p,q}\pnorm{u}{p^*,q^*}$; and (ii) for each $u$, there exists an $x$ for which $\ip{x}{u}=\pnorm{y}{p^*,q^*}$.

  Let $x$ be a vector partitioned conformally to $u$, and consider the inequality
  \begin{equation}
    \label{eq:20}
    \ip{x}{u} = \nlsum_{i=1}^g \ip{x^i}{u^i} \le \nlsum_{i=1}^g \pnorm{x^i}{q}\pnorm{u^g}{q^*},
  \end{equation}
  which follows from H\"older's inequality. Define $\psi=[\pnorm{x^i}{q}]$ and
  $\xi=[\pnorm{u^i}{q^*}]$, and invoke H\"older's inequality again to obtain
  $\ip{\psi}{\xi} \le \pnorm{\psi}{p}\pnorm{\xi}{p^*} =
  \pnorm{x}{p,q}\pnorm{u}{p^*,q^*}$. Thus, from definition~\eqref{eq.21} we
  conclude that $\pnorm{u}{*} \le \pnorm{u}{p^*,q^*}$. To prove that the dual
  norm actually equals $\pnorm{u}{p^*,q^*}$, we show that for each $u$, we can
  find an $x$ that satisfies $\pnorm{x}{p,q}=1$, for which the inner-product
  $\ip{x}{u} = \pnorm{u}{p^*,q^*}$.


  Define therefore $\beta = \sum_i \pnorm{u^i}{q^*}^{p^*}$---some juggling with indices suggests that we should set
  \begin{equation}
    \label{eq:35}
    x_j^i = \frac{1}{\beta^{1/p}}\frac{\pnorm{u^i}{q^*}^{p^*}}{\pnorm{u^i}{q^*}^{q^*}}\sgn(u^i_j)|u^i_j|^{q^*-1},
  \end{equation}
  where $x^i_j$ denotes the $j$-the element of the subvector $x^i$ (similarly $u^i_j$). 
  To see that~\eqref{eq:35} works, first consider the inner-product
  \begin{align*}
    \ip{x}{u} = \nlsum_i\ip{x^i}{u^i} &= \nlsum_i\nlsum_j x^i_ju^i_j\\
    & = \frac{1}{\beta^{1/p}}\nlsum_i \pnorm{u^i}{q^*}^{p^*-q^*}\nlsum_j |u^i_j|^{q*}\qquad
    (\text{since }\sgn(u^i_j)u^i_j=|u^i_j|)\\
    & = \frac{1}{\beta^{1/p}} \nlsum_i \pnorm{u^i}{q^*}^{p^*} = \frac{\beta}{\beta^{1/p}} = \beta^{1-1/p} = \beta^{1/p^*}\\
    & = \left(\nlsum_i \pnorm{u^i}{q^*}^{p^*}\right)^{1/p^*} = \pnorm{u}{p^*,q^*}.
  \end{align*}
  Next, we check that $\pnorm{x}{p,q}=\left(\sum_i \pnorm{x^i}{q}^p\right)^{1/p}
  = 1$. Consider thus, the term $\pnorm{x^i}{q}^p = \bigl( \sum_j  |x^i_j|^q\bigr)^{p/q}$. Using~\eqref{eq:35} we have
  \begin{align*}
    \nlsum_j |x^i_j|^q &= \frac{1}{\beta^{q/p}}\pnorm{u^i}{q^*}^{(p^*-q^*)q} \nlsum_j|u^i_j|^{q(q^*-1)}\\
    &= \frac{1}{\beta^{q/p}}\pnorm{u^i}{q^*}^{(p^*-q^*)q} \nlsum_j |u^i_j|^{q^*}\qquad&(\text{since } q^*q^{-1} +1=q^*)\\
    &= \frac{1}{\beta^{q/p}}\pnorm{u^i}{q^*}^{(p^*-q^*)q+q^*} = \frac{1}{\beta^{q/p}}\pnorm{u^i}{q^*}^{p^*q-q^*(q-1)}\\
    & = \frac{1}{\beta^{q/p}}\pnorm{u^i}{q^*}^{(p^*-1)q}&\qquad(\text{since } q(q^*)^{-1} +1=q).
  \end{align*}
  Thus, it follows that
  \begin{equation}
    \label{eq:39}
    \pnorm{x^i}{q}^p = \Bigl( \nlsum_j |x^i_j|^q\Bigr)^{p/q} = \frac{1}{\beta}\norm{u^i}{q^*}^{p(p^*-1)} = \frac{1}{\beta}\norm{u^i}{q^*}^{p^*},
  \end{equation}
  where the last equality holds because $1/p+1/p^*=1$. Finally, from~\eqref{eq:39} it follows that
  \begin{equation}
    \label{eq:41}
    \pnorm{x}{p,q}  = \bigl(\nlsum_i \pnorm{x^i}{q}^p\bigr)^{1/p} = \bigl(\tfrac{1}{\beta}\nlsum_i \pnorm{u^i}{q^*}^{p^*}\bigr)^{1/p} = 1,
  \end{equation}
  since by definition $\beta=\sum_i \pnorm{u^i}{q^*}^{p^*}$. This concludes the proof. 
\end{proof}

The next key component for Algorithm~\ref{algo2} is the proximity operator
$\prox_f$. For $f(x)=\pnorm{x}{1,q}$, this operator requires solving
\begin{equation}
  \label{eq.16}
  \min_{x^1,\ldots,x^m}\ \nlsum_{i=1}^m\half\enorm{x^i-y^i}^2 + \theta\nlsum_{i=1}^m\pnorm{x^i}{q}.
\end{equation}
Fortunately, Problem~\eqref{eq.16} separates into a sum of $m$
\emph{independent}, $\ell_q$-norm proximity operators. It suffices, therefore,
to only consider a subproblem of the form
\begin{equation}
  \label{eq.17}
  \nlmin_u\quad\half\enorm{u-v}^2 + \theta\pnorm{u}{q}.
\end{equation}
For $q=1$, the solution to~\eqref{eq.17} is given by the \emph{soft-thresholding}
operation~\cite{donoho}:
\begin{equation}
  \label{eq:10}
  u(\theta) = \sgn(v) \odot \max(|v|-\theta, 0),
\end{equation}
where operator $\odot$ performs elementwise multiplication. For
 $q=2$, we get
\begin{equation}
  \label{eq:11}
  u(\theta) = \max(1-\theta\enorm{v}^{-1}, 0)v,
\end{equation}
while the case $q=\infty$ is slightly more involved. It can be solved via the \emph{Moreau decomposition}~\cite{comPes09}, which, for a norm $f=\norm{\cdot}$ implies that
\begin{equation}
  \label{eq:12}
  \prox_{f}(v, \theta) = v - \proj_{\pol{f}}(v,\theta).
\end{equation}
For $f=\infnorm{\cdot}$, the dual-norm (polar) is $\pol{f}=\norml{\cdot}$; but
projection onto $\ell_1$-balls has been extremely well-studied---see
e.g.,~\cite{mich86,kiw07,liu09}. 

For $q > 1$ (different from $2$ and $\infty$), problem~(\ref{eq.17}) is much
harder. Fortunately, this problem was recently solved in~\cite{liu1q}, using
nested root-finding subroutines.  But unlike the cases $q \in
\set{1,2,\infty}$, the proximity operator for general $q$ can be computed only
approximately (i.e., in~(\ref{eq:18}), each iteration generates only
approximate $x(\theta)$).

\subsubsection{Mixed norms for matrices: a brief digression}
\label{sec.mtx}
We now make a brief digression, which is afforded to us by the above
results. Our digression concerns mixed-norms for matrices, as well as their
associated projection, proximity operators, which ultimately depend on
the results of the previous section. 

Our discussion is motivated by applications in~\cite{ryota}, where the
authors used mixed-norms on matrices to simultaneously. We define mixed-norms
on matrices by building upon the classic \emph{Schatten-$q$ matrix norms}~\cite{bhatia}, defined as:
\begin{equation}
  \label{eq.mtxnorm}
  \pnorm{X}{q} := \bigl(\nlsum_i \sigma_i^q(X)\bigr)^{1/q},\quad\text{for}\  q \ge 1,
\end{equation}
where $X$ is an arbitrary complex matrix, and $\sigma_i(X)$ is its $i$th
singular value. Now, let $\tx = \set{X^1,\ldots,X^m}$ be an arbitrary set of
matrices, and let $p, q \ge 1$. We define the matrix $(p,q)$-norm by the formula
\begin{equation}
  \label{eq.1}
  \pnorm{\tx}{(p,q)} := \bigl(\nlsum_{i=1}^m \pnorm{X^i}{q}^p\bigr)^{1/p}.
\end{equation}
As for the vector case, we have a similar lemma about norms dual to~\eqref{eq.1}.

\begin{lemma}[Matrix H\"older inequality]
  \label{lem.hold}
  Let $X$ and $Y$ be matrices such that $\trace(X^*Y)$ is well-defined. Then, for $p \ge 1$, such that $1/p + 1/p^* = 1$, it holds that
  \begin{equation}
    \label{eq.3}
    |\ip{X}{Y}| = |\trace(X^*Y)| \le \pnorm{X}{p}\pnorm{Y}{p^*}.
  \end{equation}
\end{lemma}
\begin{proof}
  From the well-known \emph{von Neumann} trace inequality~\cite[\S3.3]{HJ} we know that
  \begin{equation*}
    |\trace(X^*Y)| \le \nlsum_i \sigma_i(X)\sigma_i(Y) = \ip{\sigma(X)}{\sigma(Y)}.
  \end{equation*}
  Now invoke the classical H\"older inequality and use definition~\eqref{eq.mtxnorm} of matrix mixed-norms to conclude.
\end{proof}

\begin{lemma}[Dual norms]
  \label{lem.mtxdual}
  Let $p, q \ge 1$; and let $p^*, q^*$ be their conjugate exponents. The norm dual to $\pnorm{\cdot}{(p,q)}$ is $\pnorm{\cdot}{(p^*,q^*)}$.
\end{lemma}
\begin{proof}
  By the triangle-inequality and Lemma~\ref{lem.hold} we have
  \begin{equation*}
    |\ip{\tx}{\ty}|=\left|\nlsum_i \ip{X^i}{Y^i}\right| \le \nlsum_{i}|\ip{X^i}{Y^i}| \le \nlsum_i\pnorm{X^i}{q}\pnorm{Y^i}{q^*}.
  \end{equation*}
  Applying H\"older's inequality to the latter term we obtain
  \begin{equation}
    \label{eq.4}
    \nlsum_i\pnorm{X^i}{q}\pnorm{Y^i}{q^*} \le 
    \pnorm{\tx}{(p,q)}\pnorm{\ty}{(p^*,q^*)}.
  \end{equation}
  Now, we must show that for any $\ty$, we can find an $\tx$ such
  that~\eqref{eq.4} holds with equality. To that end, let $Y^i = P_iS_iQ_i^*$
  be the SVD of matrix $Y^i$. Setting $X^i = P_i\Sigma_iQ_i^*$, we see that $|\ip{\tx}{\ty}| = \nlsum_i\trace(\Sigma_iS_i)$; since both $\Sigma_i$ and $S_i$ are diagonal, this reduces to the vector case~(\ref{eq:35}), completing the proof.
\end{proof}

\noindent{\bf Projections onto $\pnorm{\cdot}{(1,q)}$-norm balls:}\\[4pt]
As for vectors, we now consider the matrix $(1,q)$-norm projection
\begin{equation}
  \label{eq.18}
  \nlmin_{X^1,\ldots,X^m}\quad\nlsum_{i=1}^m\half\frob{X^i-Y^i}^2\quad\text{s.t.}\ \nlsum_{i=1}^m\pnorm{X^i}{q} \le \gamma.
\end{equation}
Algorithm~\ref{algo2} can be used to solve~\eqref{eq.18}. The upper bound
$\theta_{\max}$ can be obtained via Lemma~\ref{lem.mtxdual}. It only remains
to solve proximity subproblems of the form
\begin{equation}
  \label{eq.14}
  \nlmin_{X}\quad\half\frob{X-Y}^2 + \theta\pnorm{X}{q}\ \ .
\end{equation}
Since both $\frob{\cdot}$ and $\pnorm{\cdot}{q}$ are unitarily invariant, from Corollary~2.5 of~\cite{lewis95} it follows that if $Y^i$ has the
singular value decomposition $Y=U\Diag(y)V^*$, then~\eqref{eq.14} is solved by
$X=U\Diag(\bar{x})V^*$, where the vector $\bar x$ is obtained by solving
\begin{equation*}
  \bar{x} := \prox_{\pnorm{\cdot}{q}}(y) := 
  \argmin\nolimits_x\quad\half\enorm{x-y}^2+\theta\pnorm{x}{q}.
\end{equation*}
We note in passing that operator~\eqref{eq.14} generalizes the popular
singular value thresholding operator~\cite{svt}, which corresponds to $q=1$
(trace norm).

\section{Algorithms for solving~\eqref{eq.9}}
\label{sec.algo}
We describe two realizations of the generic iteration~\eqref{eq.5} that can be
particularly effective: (i) spectral projected gradients; and (ii)
stochastic-gradient descent.

\subsection{Batch method: spectral projected gradient}
\label{sec.batch.algo}
The simplest method to solve~(\ref{eq.9}) is perhaps
\emph{gradient-projection}~\cite{gp}, where starting with a suitable initial
point $x_0$, one iterates
\begin{equation}
  \label{eq:9}
  x_{t+1} = \proj_{f}(x_t-\eta_t\nabla L(x_t)),\quad t=0,1,\ldots.
\end{equation}
We have already discussed $\proj_f$; the other two important parts
of~\eqref{eq:9} are the stepsize $\eta_t$, and the gradient $\nabla L$. Even
when the loss $L$ is not convex, under fairly mild condition, we may still
iterate~\eqref{eq:9} to obtain convergence to a stationary
point---see~\cite[Chapter 1]{bertsekas} for a detailed discussion, including
various strategies for computing stepsizes. If, however, $L$ is convex, we may
invoke a method that typically converges much faster: \emph{spectral projected
  gradient} (SPG)~\cite{spg}.

SPG extends ordinary gradient-projection by using the famous (nonmonotonic) \emph{spectral stepsizes} of Barzilai and Borwein~\cite{bb} (BB). Formally, these stepsizes are
\begin{equation}
  \label{eq:19}
  \eta_{BB1} := \frac{\ip{\Delta x_t}{\Delta x_t}}{\ip{\Delta g_t}{\Delta x_t}}, 
  \quad\text{or}\quad
  \eta_{BB2} := \frac{\ip{\Delta x_t}{\Delta g_t}}{\ip{\Delta g_t}{\Delta g_t}},
\end{equation}
where $\Delta x_t = x_t - x_{t-1}$, and $\Delta g_t = \nabla L(x_t)-\nabla L(x_{t-1})$. 

SPG substitutes stepsizes~\eqref{eq:19} in~\eqref{eq:9} (using
safeguards to ensure bounded steps). Thereby, it leverages the
strong empirical performance enjoyed by BB
stepsizes~\cite{bb,spg,dai,markpqn}; to ensure global convergence, SPG invokes
a nonmontone line search strategy that allows the objective value to
occasionally increase, while maintaining some information that allows
extraction of a descending subsequence.

\paragraph{Inexact projections:} Theoretically, the convergence analysis of
SPG~\cite{spg} depends on access to a subroutine that computes $\proj_f$
\emph{exactly}. Obviously, in general, this operator cannot be computed
exactly (including for many of the mixed-norms). To be correct, we must rely
on an \emph{inexact} SPG method such as~\cite{spg.inex}. In fact, due to
roundoff error, even the so-called exact methods run inexactly. So, to be
fully correct, we must treat the entire iteration~\eqref{eq:9} as being
inexact. Such analysis can be done (see~e.g.,~\cite{polyak}); but it is not
one of the main aims of this paper, so we omit it.

\subsection{Stochastic-gradient method}
\label{sec.online}
Suppose the loss-function $L$ in~\eqref{eq.9} is separable, that is,
\begin{equation}
  \label{eq.12}
  L(x) = \nlsum_{i=1}^r \ell_i(x),\qquad\text{where}\ x \in \reals^d,
\end{equation}
for some large number $r$ of components (say $r \gg d$). In such a case,
computing the entire gradient $\nabla L$ at each iteration~(\ref{eq:9}) may be
too expensive, and it might be more preferable to use stochastic-gradient
descent (SGD)\footnote{This popular name is a misnomer because SGD does
  \emph{not} necessarily lead to descent at each step.} instead. In its
simplest realization, at iteration $t$, SGD picks a random index $s(t) \in
[r]$, and replaces $\nabla L(x)$ by a stochastic estimate $\nabla
\ell_{s(t)}(x)$. This results in the iteration
\begin{equation}
  \label{eq.13}
  x_{t+1} = \proj_{f}(x_t - \eta_t\nabla\ell_{s(t)}(x_t)),\quad t=0,1,\ldots,
\end{equation}
where $\eta_t$ are suitable (e.g., $\eta_t \propto 1/t$) stepsizes. Again,
some additional analysis is also needed for~\eqref{eq.13} to account for the
potential inexactness of the projections.

\vspace*{-12pt}
\section{Experimental results and applications}
\label{sec.results}
We present below numerical results that illustrate the computational
performance of our methods. In particular, we show the following main
experiments:
\begin{enumerate}
\item Running time behavior of our root-finding projection methods, including
  \begin{itemize}
  \item Comparisons against the method of~\cite{darell} for $\ell_{1,\infty}$ projections
  \item Some results on $\ell_{1,q}$ projections for a few different values of $q$.
  \end{itemize}
\item Application to the $\ell_{1,\infty}$-norm multitask lasso~\cite{liPaZh09}, for which we show
  \begin{itemize}
  \item Running time behavior of SPG, both with our projection and that of~\cite{darell};
  \item Derivation of and numerical results with a SGD based method for MTL.
  \end{itemize}
\end{enumerate}

\vspace*{-12pt}
\subsection{Projection onto the $\ell_{1,\infty}$-ball}
\vspace*{-8pt}
For ease of comparison, we use the notation of~\cite{darell}, who seem to be the first to consider efficient projections onto the $\oneinf$-norm ball. The task is to solve
\begin{equation}
  \label{eq:7}
  \nlmin_{W}\quad\half\frob{W-V}^2,\quad\text{s.t.}\quad\nlsum_{i=1}^d\infnorm{w^i} \le \gamma,
\end{equation}
where $W$ is a $d \times n$ matrix, and $w^i$ denotes its $i$th row.

In our comparisons, we refer to the algorithm of~\cite{darell} (C
implementation)\footnote{\textit{http://www.lsi.upc.edu/$\sim$aquattoni/CodeToShare/}},
as `QP',\footnote{The runtimes for QP reported in this paper differ
  significantly from those in our previous paper~\cite{ecml11}. This
  difference is due to an unfortunate bug in the previous implementation
  of~\cite{darell}, which got uncovered after the authors of~\cite{darell} saw
  our experimental results in~\cite{ecml11}.} and to our method as `FP' (also
C implementation). The experiments were run on a single core of a quad-core AMD Opteron (2.6GHz), 64bit Linux machine with 16GB RAM.

We compute the optimal $W^*$, as $\gamma$ varies from $0.01\oneinfnorm{V}$
(more sparse) to $0.6\oneinfnorm{V}$ (less sparse)
settings. Tables~\ref{tab.1}--\ref{tab.3} present running times, objective
function values, and errors (as measured by the constraint violation:
$|\gamma-\oneinfnorm{W^*}|$, for an estimated $W^*$). The tables also show the
absolute difference in objective value between QP and FP. While for small
problems, QP is very competitive, for larger ones, FP consistently outperforms
it. Although on average FP is only about twice as fast as QP, it is noteworthy
that despite FP being an ``inexact'' method (and QP an ``exact'' one), FP
obtains solutions of accuracy many magnitudes of order better than QP.

\begin{table}[tbp]\scriptsize
  \centering
  \begin{tabular}{r|r|r||r|r||r|}
    $\frac{\gamma}{\oneinfnorm{V}}$ & QP$_\text{time}$ (s)& FP$_\text{time}$ (s) &
    QP$_\text{err}$ & FP$_\text{err}$ & $|$FP$_\text{obj}$-QP$_\text{obj}|$\\
    \hline
    0.01 &21.90 &11.57 &3.17E-06 &5.12E-13 &1.36E-06\\
    0.05 &22.23 &11.70 &2.61E-06 &4.55E-13 &1.04E-06\\
    0.10 &21.60 &12.71 &2.00E-06 &4.55E-13 &7.22E-07\\
    0.20 &20.71 &14.33 &1.10E-06 &1.82E-12 &3.13E-07\\
    0.30 &19.87 &14.33 &5.51E-07 &0.00E+00 &1.18E-07\\
    0.40 &19.64 &18.36 &2.48E-07 &1.82E-12 &3.76E-08\\
    0.50 &19.21 &16.50 &9.82E-08 &0.00E+00 &9.98E-09\\
    0.60 &19.04 &17.09 &3.33E-08 &0.00E+00 &2.15E-09\\
    \hline
  \end{tabular}
  \caption{\small Runtime and accuracy for QP and FP  on a $10,000\times 300$ matrix $V$.}
  \label{tab.1}
\end{table}

\begin{table}[tbp]\scriptsize
  \centering
  \begin{tabular}{r|r|r||r|r||r|}
    $\frac{\gamma}{\oneinfnorm{V}}$ & QP$_\text{time}$ (s)& FP$_\text{time}$ (s) &
    QP$_\text{err}$ & FP$_\text{err}$ & $|$FP$_\text{obj}$-QP$_\text{obj}|$\\
    \hline
    0.01 &38.08 &22.00 &1.05E-05 &7.28E-12 &1.16E-06\\
    0.05 &39.30 &20.86 &8.74E-06 &1.82E-12 &9.08E-07\\
    0.10 &39.27 &21.19 &6.88E-06 &3.64E-12 &6.53E-07\\
    0.20 &38.51 &23.94 &4.04E-06 &7.28E-12 &3.09E-07\\
    0.30 &38.27 &24.07 &2.20E-06 &2.18E-11 &1.30E-07\\
    0.40 &37.92 &31.10 &1.12E-06 &1.46E-11 &4.91E-08\\
    0.50 &39.40 &27.82 &5.22E-07 &0.00E+00 &1.61E-08\\
    0.60 &37.47 &27.36 &2.16E-07 &0.00E+00 &4.54E-09\\
    \hline
  \end{tabular}
  \caption{\small Runtime and accuracy for QP and FP  on a $50,000\times 1000$ matrix $V$.}
  \label{tab.2}
\end{table}

\begin{table}[tbp]\scriptsize
  \centering
  \begin{tabular}{r|r|r||r|r||r|}
    $\frac{\gamma}{\oneinfnorm{V}}$ & QP$_\text{time}$ (s)& FP$_\text{time}$ (s) &
    QP$_\text{err}$ & FP$_\text{err}$ & $|$FP$_\text{obj}$-QP$_\text{obj}|$\\
    \hline
    0.01 &521.13 &187.61 &1.21E-04 &1.14E-12 &4.24E-05\\
    0.05 &528.00 &197.96 &9.78E-05 &1.82E-12 &3.17E-05\\
    0.10 &526.18 &228.55 &7.33E-05 &3.64E-12 &2.13E-05\\
    0.20 &492.04 &257.08 &3.81E-05 &1.46E-11 &8.50E-06\\
    0.30 &466.76 &256.54 &1.77E-05 &1.46E-11 &2.86E-06\\
    0.40 &454.75 &247.34 &7.33E-06 &0.00E+00 &8.06E-07\\
    0.50 &447.80 &305.13 &2.71E-06 &1.46E-11 &1.90E-07\\
    0.60 &444.56 &236.83 &8.73E-07 &0.00E+00 &3.63E-08\\
    \hline
  \end{tabular}
  \caption{\small Runtime and accuracy for QP and FP  on a $50,000\times 10,000$ matrix $V$. For this experiment, QP did not run on our machine with 16GB, so we performed this experiment on a machine with 32GB RAM.}
  \label{tab.3}
\end{table}

\subsection{Projection onto $\ell_{1,q}$-balls}
\begin{figure}[htbp]\centering
  \begin{tabular}{cc}
    \hskip -1cm\includegraphics[scale=.2]{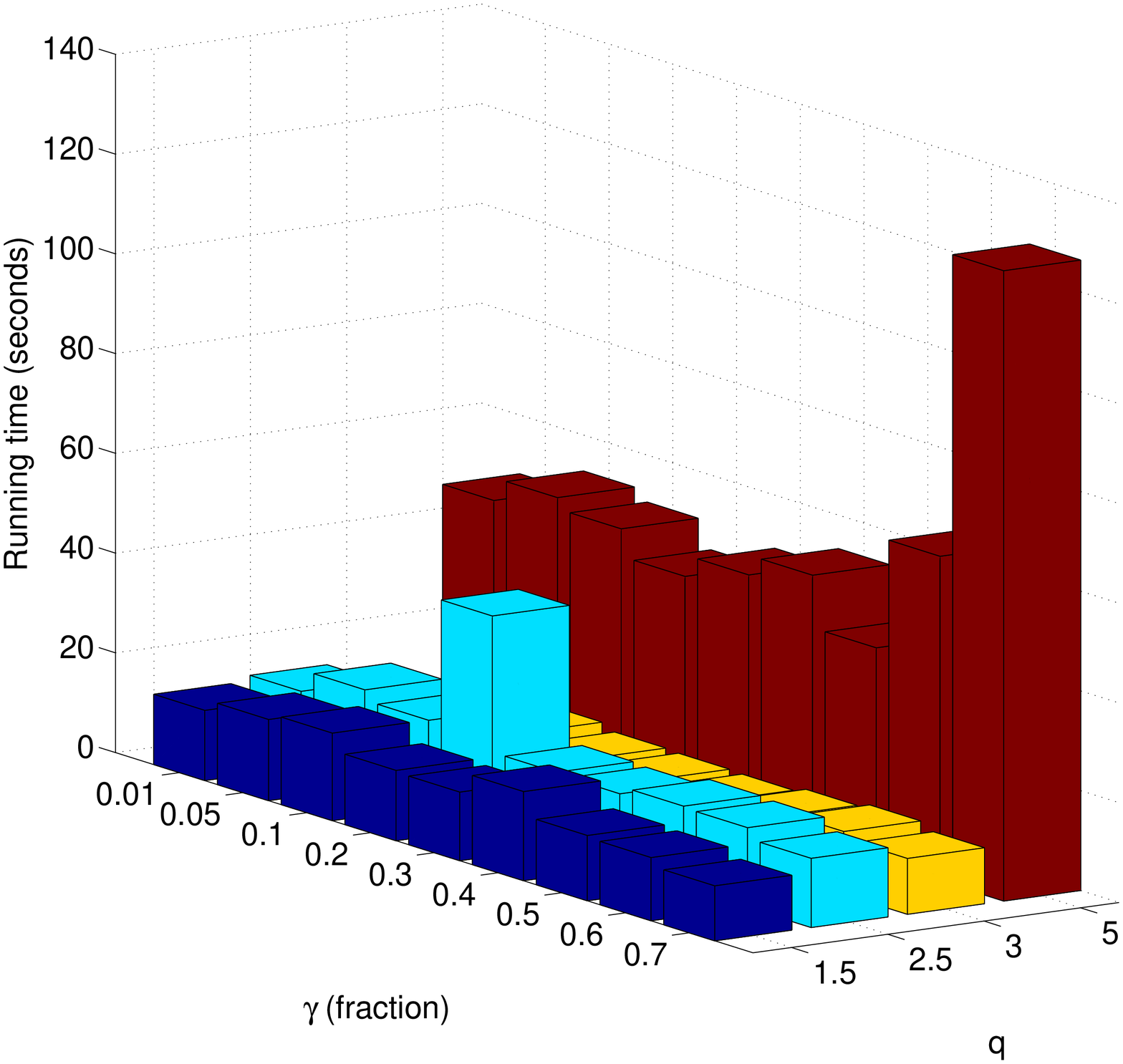} &\hskip -1cm
    \includegraphics[scale=.2]{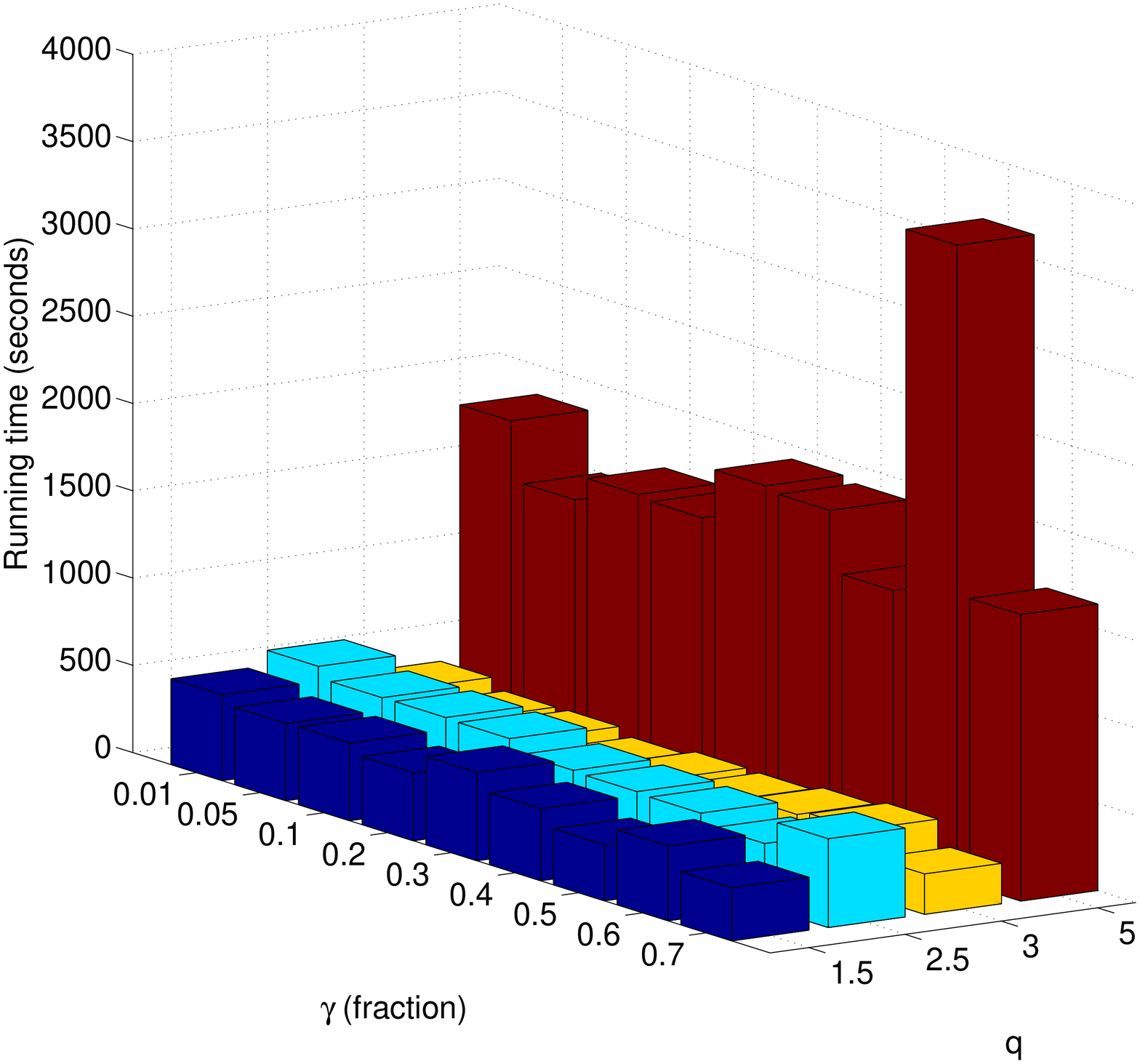}
  \end{tabular}
  \caption{Running times for $\ell_{1,q}$-norm projections as scalars $q$ and ratios $\gamma/\pnorm{V}{1,q}$ vary. The left plot is on a ${1000 \times 100}$ matrix, while the right one is on a $5000 \times 500$ matrix.}
\label{fig:l1q}
\end{figure}

Next we show running time behavior displayed our method for projecting onto $\ell_{1,q}$ balls; we show results for $q \in \set{1.5, 2.5, 3, 5}$, when solving 
\begin{equation}
  \label{eq.22}
  \nlmin_{W}\quad\frob{W-V}^2,\quad\text{s.t.}\ \ \nlsum_{i=1}^d \pnorm{w^i}{q}.
\end{equation}
The plots (Figure~\ref{fig:l1q}) also running time behavior as the parameter
$\gamma$ is varied. These plots reveal four main points: (i) the runtimes seem
to be largely independent of $\gamma$; (ii) for smaller values of $q$, the
projection times are approximately same; and (iii) for larger values of, the
projection times increase dramatically. 

Moreover, from the actual running times it is apparent our projection code scales linearly with the data size. For example, the matrix corresponding to the second bar plot has 25 times more parameters than the first plot, and the runtimes reported in the second plot are approximately 25--30 times higher. Although the running times scale linearly, a single $\ell_{1,q}$-norm projection still takes
nontrivial effort. Thus, even though our $\ell_{1,q}$-projection method is
relatively fast, currently we can recommend it only for small and medium-scale
regression problems.

\subsection{Application to Multitask Lasso}
\label{sec:appl}
Multitask Lasso~(\mtl)~\cite{tuVeWr05,liPaZh09} is a simple grouped feature
selection problem, which separates important features from less important ones
by using information shared across multiple tasks. The feature selection is
effected by a sparsity promoting mixed-norm, usually the
$\ell_{1,\infty}$-norm~\cite{liPaZh09}.

Formally, \mtl is setup as follows. Let $\mx_j \in \reals^{m_j \times d}$ be
the data matrix for task $j$, where \fromto{1}{j}{n}. \mtl seeks a matrix $W
\in \reals^{d \times n}$, each column of which corresponds to parameters for a
task; these parameters are regularized across features by applying a
mixed-norm over the rows of $W$. This leads to a ``grouped'' feature
selection, because if for a row, the norm $\infnorm{w^i}=0$, then the entire
row $w^i$ gets eliminated (i.e., feature $i$ is removed). The standard \mtl
optimization problem is
\begin{equation}
  \label{eq:mtl}
  \min_{w_1,\ldots,w_n}\quad \Lc(W) := \nlsum_{j=1}^n 
  \half\enorm{y_j - X_jw_j}^2,\quad\text{s.t.}\quad
  \nlsum_{i=1}^d\pnorm{w^i}{\infty} \le \gamma,
\end{equation}
where the $y_j$ are the dependent variables, and $\gamma > 0$ is a
sparsity-tuning parameter. Notice that the loss-function combines the
different tasks (over columns of $W$), but the overall problem does not
decompose into separable problems because the mixed-norm constrained is over
the \emph{rows} of $W$.

\subsubsection{Stochastic-gradient based MTL}
We may rewrite the MTL problem as
\begin{equation}
  \label{eq.10}
  \begin{split}
    \min\ L(W) := &\nlsum_{j=1}^n\half\enorm{y_j-X_jw_j}^2 =  \half\enorm{y-Xw}^2,\\
    \text{s.t.}\quad&\nlsum_{i=1}^d\pnorm{w^i}{\infty} \le \gamma,
  \end{split}
\end{equation}
where we have introduced the notation
\begin{equation*}
  y = \vect(Y),\quad X=X_1\oplus \cdots \oplus X_n,\quad\text{and}\ w = \vect(W),
\end{equation*}
in which $\vect(\cdot)$ is the operator that stacks columns of its argument to yield a long vector, and $\oplus$ denotes the direct sum of two matrices. Notice that if it were not for the $\ell_{1,\infty}$-norm constraint, problem~\eqref{eq.10} would just reduce to ordinary least squares. 

The form~\eqref{eq.10}, however, makes it apparent how to derive a stochastic-gradient method. In particular, suppose that we use a ``mini-batch'' of size $b$, i.e., we choose $b$ rows of matrix $X$, say $X_b$. Let $y_b$ denote the corresponding rows (components) of $y$. This subset of rows contributes $\ell_b(w) := \half\enorm{y_b-X_bw}^2$ to the objective~\eqref{eq.10}, whereby we have the stochastic-gradient
\begin{equation}
  \label{eq.24}
  \nabla\ell_b(w) = X_b^T(X_bw-y_b).
\end{equation}
Then, upon instantiating iteration~(\ref{eq.13}) with~\eqref{eq.24}, we obtain  Algorithm~\ref{mtlsg}.
\begin{algorithm2e}[tbp]
  \KwIn{Scalar $\gamma > 0$; batchsize $b$; stepsize sequence: $\eta_0,\eta_1,\ldots$}
  \KwOut{$W^* \approx \argmin_{W}\ L(W)$, s.t.\ $\pnorm{W^T}{1,\infty} \le \gamma$}
  $W_0 \gets 0$\;
  \While{$\neg$ converged}{
    Pick $b$ different indices in $[mn]$\;
    Obtain stochastic gradient using~\eqref{eq.24}\;
    $W_t \gets \proj(W_{t-1} - \eta_t\nabla\ell_s(W_t))$\;
    $t \gets t + 1$\;
  }
  \KwRet{$W^*$}
 \caption{\small MTL via stochastic-gradient descent} \label{mtlsg}
\end{algorithm2e}

\paragraph{Implementation notes:} Despite our careful implementation, for
large-scale problems the projection can become the bottleneck in
Algorithm~\ref{mtlsg}. To counter this, we should perform projections only
occasionally---the convergence analysis is unimpeded, as we may restrict our
attention to the subsequence of iterates for which projection was
performed. Other implementation choices such as size of the mini-batch and the
values of the stepsizes $\eta_t$ are best determined empirically. Although
tuning $\eta_t$ can be difficult, this drawback is offset by the gain in
scalability.

\subsubsection{Simulation results}
We illustrate running time results of SPG on two large-scale instances of \mtl
(see Table~\ref{tab:data}). We report running time comparisons between two
different invocations of an SPG-based method for solving~\eqref{eq:mtl}, once
with QP as the projection method and once with FP---we call the corresponding
solvers SPG$_{\text{QP}}$, and SPG$_{\text{FP}}$. We note in passing that
other efficient \mtl algorithms (e.g.,~\cite{icml10,liu1q}) solve the
\emph{penalized} version; our \emph{formulation} is constrained, so we only
show SPG.

\begin{table}[h]\small
  \centering
  \begin{tabular}{c||l|l}
    Name & $(m, d, n)$ & \#nonzeros\\
    \hline
    D1 & (1K, 5K,  10K) & 50 million\\
    D2 & (10K, 50K, 1K) & 500 million
  \end{tabular}
  \caption{\small Sparse datasets used for MTL. For simplicity, all matrices $X_j$ (for each task $1\le j\le n$), were chosen to have the same size ${m \times d}$.}
  \label{tab:data}
\end{table}

\begin{table}[htbp]\small
  \centering
  \begin{tabular}{c|r||r|r||r|r}
     Dataset &\#projs & proj$_{\text{QP}}$ & proj$_{\text{FP}}$ & SPG$_{\text{QP}}$ & SPG$_{\text{FP}}$\\
    \hline
    D1 & 50   &   2275.2s & \tb{1204.3s} & 2722.9s & \tb{1728.3s}\\
    D2 & 48   &   2631.8s & \tb{1362.3s} & 3495.1s & \tb{2296.7s}
  \end{tabular}
  \caption{\small Running times (seconds) on datasets D1 and D2. SPG was used to solve MTL, with stopping tolerance of $10^{-5}$. Total number of projections required to reach this accuracy are reported as '\#projs'. The columns 'proj$_{\text{QP}}$' and 'proj$_{\text{FP}}$', report the total time spent by the \spgqp and \spgfp methods for the $\oneinf$-projections alone. The last two columns report the overall time taken by \spgqp and \spgfp.}
  \label{tab:mtl}
\end{table}
The results in Table~\ref{tab:data} indicate that for large-scale problems,
the savings accrued upon using our faster projections (in combination with
SPG) can be substantial.

\subsubsection{MTL results on real-world data}
We now show a running comparison between three methods: (i) \spgqp, (ii)
\spgfp, and (iii) SGD (with projection step computed using FP). For our
comparison, we solve \mtl on a subset of the CMU Newsgroups
dataset\footnote{Original at:
  \textit{http://www.cs.cmu.edu/$\sim$textlearning/}; we use the reduced version of~\cite{icml10}.}. 

The dataset corresponds to 5 feature selection tasks based on data taken from
the following newsgroups: computer, politics, science, recreation, and
religion. The feature selection tasks are spread over the matrices
$\mx_1,\ldots,\mx_5$, each of size $2907 \times 53975$, while the dependent
variables $\vy_1,\ldots,\vy_5$ correspond to class labels.

\begin{figure}[ht]
  \centering
  \includegraphics[height=4cm,width=.45\linewidth]{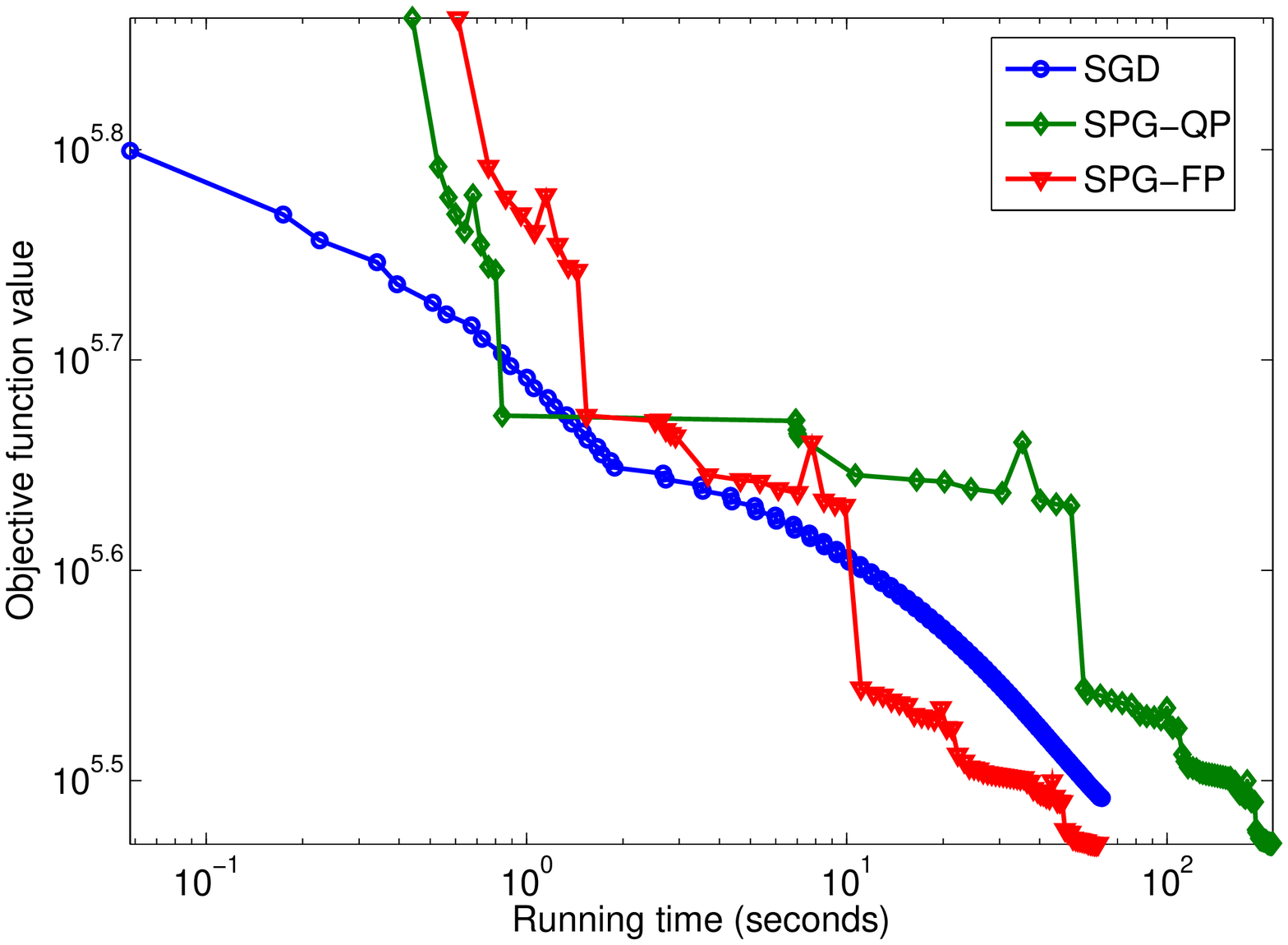}
  \includegraphics[height=4cm,width=.45\linewidth]{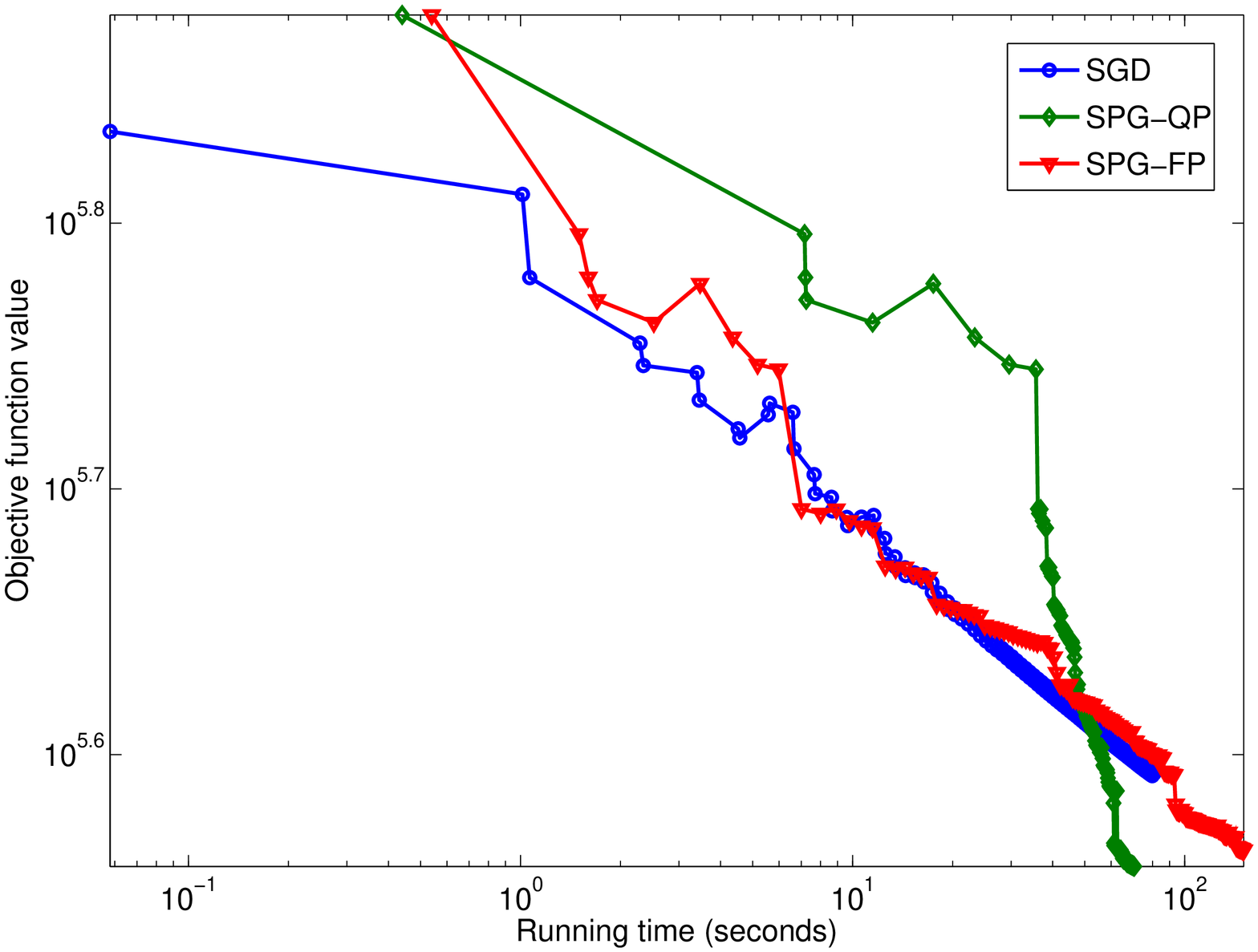}
  \caption{Running time results on CMU Newsgroups subset (left: less sparse; right: more sparse problem).}
  \label{fig.cmu}
\end{figure}

Figure~\ref{fig.cmu} reports running time results obtained by the three
methods in question (all methods were initialized by the same $W_0$). As
expected, the stochastic-gradient based method rapidly achieves a low-accuracy
solution, but start slowing down as time proceeds, and eventually gets
overtaken by the SPG based methods. Interestingly, in the first experiment,
\spgqp takes much longer than \spgfp to convergence, while in the second
experiment, it lags behind substantially before accelerating towards the
end. We attribute this difference to the difficulty of the projection
subproblem: in the beginning, the sparsity pattern has not yet emerged, which
drives \spgqp to take more time. In general, however, from the figure it seems
that either SGD or \spgfp yield an approximate solution more rapidly---so for
problems of increasingly larger size, we might prefer them.\footnote{Though some effort must always be spent to tune the batch and stepsizes for SGD.}

\section{Discussion}
\label{sec.disc}
We described mixed-norms for vectors, which we then naturally extended also to
matrices. We presented some duality theory, which enabled us to derive
root-finding algorithms for efficiently computing projections onto mixed-norm
balls, especially for the special class of $\ell_{1,q}$-mixed norms. For
solving an overall regression problem involving mixed-norms we suggested two
main algorithms, spectral projected gradient and stochastic-gradient (for
separable losses). We presented a small but indicative set of experiments to
illustrate the computational benefits of our ideas, in particular for the
multitask lasso problem.

At this point, several directions of future work remain open---for instance:
\begin{itemize}
\item Designing fast projection methods for certain classes of non-separable
  mixed norms. Some algorithms already exist for particular classes~\cite{bach,bach2}.
\item Studying norm projections with additional simple constraints (e.g., bounds).
\item Extending the fast methods of this paper to non-Euclidean proximity operators.
\item Exploring applications of matrix mixed-norm regularizers.
\end{itemize}

\bibliographystyle{spmpsci}      

\end{document}